\documentclass{article}
\usepackage{ijcai24}

\usepackage{authblk}
\usepackage{times}
\usepackage{soul}
\usepackage{url}
\usepackage[hidelinks]{hyperref}
\usepackage[utf8]{inputenc}
\usepackage[small]{caption}
\usepackage{graphicx}
\usepackage{amsmath}
\usepackage{amsthm}
\usepackage{booktabs}
\usepackage{algorithm}
\usepackage[switch]{lineno}

\usepackage{boldline} 
\usepackage{multirow}
\usepackage{sidecap}
\usepackage{subcaption}
\usepackage{amsmath}
\usepackage{amsfonts}
\usepackage{xcolor}
\usepackage{bm}
\usepackage{hyperref}
\usepackage{rotating}
\usepackage{adjustbox}
\usepackage{algorithm}
\usepackage{algpseudocode} 
\usepackage{array}
\definecolor{green}{rgb}{0.0, 0.5, 0.0}
\definecolor{red}{rgb}{0.82, 0.1, 0.26}
\definecolor{codegray}{gray}{0.9}

\usepackage{pifont}
\newcommand{\cmark}{\ding{51}}%
\newcommand{\xmark}{\ding{55}}%

\newlength{\Oldarrayrulewidth}

\title{ Rethinking ChatGPT's Success: Usability and Cognitive Behaviors Enabled by Auto-regressive LLMs' Prompting }

\author{\textbf{Xinzhe Li}, \textbf{Ming Liu}}
\affil{School of IT, Deakin University, Australia \\
\texttt{\{lixinzhe, m.liu\}@deakin.edu.au}}

\begin{document}

\maketitle

\begin{abstract}
Over the last decade, a wide range of training and deployment strategies for Large Language Models (LLMs) have emerged. Among these, the prompting paradigms of Auto-regressive LLMs (AR-LLMs) have catalyzed a significant surge in Artificial Intelligence (AI). This paper aims to emphasize \emph{the significance of utilizing free-form \footnote{``Free-form'' describes a stream of meaningful symbols, created by humans or through auto-regressive methods.} modalities (forms of
input and output) and verbal free-form contexts as user-directed channels (methods for transforming modalities) for downstream deployment}.
Specifically, we analyze the structure of modalities within both two types of LLMs and six task-specific channels during deployment. From the perspective of users, our analysis introduces and applies the analytical metrics of task customizability, transparency, and complexity to gauge their usability, highlighting the superior nature of AR-LLMs' prompting paradigms.
Moreover, we examine the stimulation of diverse cognitive behaviors in LLMs through the adoption of free-form text and verbal contexts, mirroring human linguistic expressions of such behaviors.
We then detail four common cognitive behaviors to underscore how AR-LLMs' prompting successfully imitate human-like behaviors using this free-form modality and channel.
Lastly, the potential for improving LLM deployment, both as autonomous agents and within multi-agent systems, is identified via cognitive behavior concepts and principles.
\end{abstract}

\section{Introduction}
ChatGPT has emerged as the most popular AI application, with a vast user base. The success of GPT models can be attributed to the scaling of transformer-based neural networks and the extensive pre-training data, as explored in previous studies \citep{radford2019language,brown2020gpt3}. 
The scope of this paper is directed towards Large Language Models (LLMs) that are sufficiently large to acquire world knowledge, commonsense, and the linguistic capabilities required to attain high performance on benchmarks such as GLUE \citep{wang2018glue}.

Although LLMs are commonly perceived as general-purpose language intelligence models, the practice often diverges from employing a singular, all-encompassing model for every task. Instead, the deployment frequently entails developing a suite of specialized models tailored to specific tasks. This specialization is facilitated through the introduction of task-specific channels, modifying the model's structure or its pre-trained parameters to better suit the nuances of individual tasks. This highlights a departure from the ideal of a universal, one-size-fits-all model, while the broad capabilities of LLMs suggest they could serve as jack-of-all-trades in language processing.
This trend towards creating task-specific models may stem from the tradition of evaluating linguistic intelligence through a variety of distinct tasks and benchmarks \citep{wang2018glue}, with researchers striving to excel in these tasks independently to set new benchmarks. In this paper, we delve into the mechanisms behind prevalent deployment paradigms including AR-LLMs' prompting, which underpins ChatGPT's operation, and highlight several critical observations:
1) Models tailored with optimized task-specific channels often suffer from issues related to task customizability, transparency, and user-level complexity during deployment, affecting their overall usability;
2) Anticipated to mimic human-like intelligence, they often exhibit slow thinking through shortcuts \citep{kahneman2011thinking};
3) They frequently fall short in showcasing advanced cognitive behaviors, which we contend are vital for convincing users of the models' intelligence. Conversely, AR-LLMs' prompting paradigms introduce a more natural, human-like channel (verbal free-form context) for representing a wide array of real-life tasks and employ form-form output modalities to showcase cognitive behaviors in complex scenarios.

Specifically, in this paper, we commence by examining the foundational principles of language modeling, revisiting the notable split in language modeling approaches that emerged in the late 2010s: auto-encoding LMs (AE-LMs) exemplified by BERT \citep{jin2020bert} and auto-regressive LMs (AR-LMs) exemplified by the GPT series \citep{radford2018improving,brown2020gpt3}. Rather than delve into an extensive array of deployment paradigms, we introduce and discuss the concepts of modalities and channels to investigate how LLMs are deployed (\S \ref{sec:modality_channel}).
Upon evaluating different deployment paradigms for LLMs, it becomes clear that aside from the AR-LLMs' prompting approach, other paradigms struggle to demonstrate advanced human-like cognitive behaviors. This shortfall is attributed to the constraints within modalities and channels, coupled with a tendency towards superficial learning, i.e., slow thinking (\S \ref{sec:comparisons_of_paradigms} and \S \ref{sec:think_fast_slow}).
In contrast, via specified context in the free-from text, the AR-LLMs' prompting strategy imitate human-like cognitive behaviors, such as reasoning, planning, and feedback learning, which are elucidated in Table \ref{tab:think_plan_act} (\S \ref{sec:think_plan_act}).
Finally, we explore how understanding cognitive behaviors can help overcome the tuning and deployment obstacles encountered by LLMs functioning as autonomous entities and within multi-agent frameworks (\S \ref{sec:challenges}).

\section{Deploying Large Language Models}
\label{sec:modality_channel}
This section elucidates the dual objectives underlying language models, which both aim to model the joint probability distribution of text sequences through self-supervised learning techniques and generate text that is relevant to the given context. 
After this introduction, we present a novel framework that facilitate the characterization of various deployment paradigms through two types of data modalities, which support language comprehension, coupled with six unique channels for processing these modalities.

\subsection{The Fundamental Dichotomy in Language Modeling} 
\label{sec:dichotomy}
\paragraph{Objective of Language Modeling}
The goal of language modeling is to estimate the joint probability distribution of sequences of text \citep{bengio-2003-neural-lm}. This involves developing two distinct yet relaxed formulations for constructing LLMs that leverage self-supervised learning from vast quantities of unlabeled text data. The self-supervised approach enables the training of LLMs on extensive text corpora, a practice that has been thoroughly investigated in various studies \citep{liu2019roberta,wei2022emergent}. This paper focuses on how the intrinsic design of language models impacts their usability and potential to express cognitive behaviors.

\paragraph{Auto-Regressive (Left-to-Right) Language Modeling}
Typically, language modeling is approached by predicting the subsequent token in a sequence based on the preceding tokens. This prediction is quantified as the product of conditional probabilities for each subsequent token, considering its previous tokens, in accordance with the chain rule \citep{bengio-2003-neural-lm}.
\begin{equation} \label{eq:left_to_right_lm}
\begin{split}
P\left( w_{1}, \ldots, w_{N} \right) = \prod_{t=1}^N P\left(w_{t} \mid w_0, \ldots, w_{t-1} \right)
\end{split}
\end{equation}
Here, $w_0$ serves as a marker for the beginning of text.

\paragraph{Auto-Encoding (Denoising) Language Modeling}
In the context of auto-encoding language modeling, noise is intentionally introduced to an input sequence \( w_1, w_2, ... w_N \). The primary aim is to optimize
\begin{equation} \label{eq:denoising_lm}
\max  \prod_{t=1}^N  P\left(w_{t} \mid \hat{w}_1, \ldots, \hat{w}_{N} \right)
\end{equation}
where \( \hat{w}_1, \hat{w}_2, ... \hat{w}_N \) represents the altered, noise-added version of the input sequence.
The approach of masking specific tokens in the text at random, known as token-level masked language modeling \citep{devlin-bert2018}, is a widely adopted strategy. This involves substituting original tokens with a special token, such as ``[MASK]'', and training the model to predict these original tokens based on the context of the surrounding, unmasked tokens. The discrepancy between the original and reconstructed sequences is quantified through a reconstruction loss:
\begin{equation} \label{eq:denoising_lm2}
L_{reconstruction} = -\sum_{t=1}^N \log P\left(w_{t} \mid \hat{w}_1, \ldots, \hat{w}_{N} \right)
\end{equation}
This denoising methodology also includes other variants such as span-level masked language modeling \citep{joshi-etal-2020-spanbert}, text infilling \citep{lewis-etal-2020-bart}, among others.

\subsection{Exploring the Modalities within Large Language Models}
This section delves into the concept of ``modalities'' within LLMs, a term often implicitly associated with research on multimodal systems to describe diverse, human-like channels of communication, such as text, speech, gestures, and visual inputs \citep{bartneck2020human}. Here, ``modalities'' specifically refer to the various forms of input and output data utilized in LLM deployment.

In the operation of both AR-LLMs and AE-LLMs, we identify three primary modalities: a unique textual modality for both the input and output in AR-LLMs (unrestricted text), a distinct textual modality for AE-LLMs (masked text or contextualized n-grams), and a shared modality of intermediate dense representations applicable to both models:
1) \textbf{Intermediate Dense Representations}: Fundamentally, LLMs convert each word (or subword) in a sequence into dense vector embeddings. These embeddings are generated through a series of mathematical operations, such as the self-attention mechanism, at every layer of the neural network, and are represented as $\left\{h_i^l\right\}$ for every position $i$ within the sequence and for every layer $l$ in the model. Here, $i$ ranges from $1$ to $N$, with $N$ indicating the total number of elements in the sequence, and $l$ spans from $1$ to $L$, where $L$ represents the complete count of layers within the model.
2) \textbf{Textual Modalities}: AE-LLMs feature an input modality of masked text, with the output modality being contextualized n-grams designed to reconstruct the masked sections. Conversely, due to their auto-regressive design, AR-LLMs are capable of encoding any text as context and generating free-form text outputs, thereby employing unrestricted text for both input and output. These modalities are inherently linked to their respective language modeling strategies. 

\begin{table*}[ht!]
    \centering
    \small
    \begin{tabular}{lp{3cm}p{2cm}p{2.5cm}p{2cm}}
    \toprule 
       Channels   & Relevant Paradigms & Customizability & Transparency & Complexity \\
    \midrule

       Adapter  &   Adapter tuning & \xmark & \xmark & $T$
       \\ \hline
       
       Output layers & LLM fine-tuning; Adapter tuning & \xmark & \xmark & $T$
       \\\hline
       
       LLMs & LLM fine-tuning; PET & \xmark & \xmark 
       & $T$
       \\ \hline
       
       Activation prefixes & Prefix tuning & \xmark & \xmark 
       &  $T$
       \\ \hline
       
       Verbal free-form context & AR-LLMs' prompting & \cmark & \cmark 
       & $0$
       \\ \hline
       
       Contextual text patterns & PET; \newline Auto-prompt & \xmark & \cmark (PET); \newline \xmark (Auto-prompt) 
       & $N \times T$
       \\ \bottomrule
    \end{tabular}
    \caption{Evaluation of deployment channels for language models: A comparative analysis of task customizability, transparency and complexity from the users' perspective. PET: Pattern exploitation training; $T$: the total number of task; $N$: the number of patterns per task.}
    \label{tab:assess_channel}
\end{table*}

\subsection{Task-specific Channels for Deployment}
To tailor the core capabilities of LLMs for specific downstream tasks, both input and intermediate modalities can be altered directly (for instance, by appending prefixes or incorporating verbal context) or indirectly through the use of parametric modules such as neural networks, including adapters and output layers as described subsequently. It's worth noting that direct modifications, such as prefixes, can also be achieved using parametric modules. These parametric modules undergo optimization via task-specific supervised learning.  In this context, we describe the means for modality transformation aimed at specific tasks as task-specific channels. 
For clarify, modalities are the types of data or the form in which data is processed, while channels are the pathways or methods through which these data modalities are adapted or transformed for specific tasks. Task-specific channels encompass:
1) \textbf{Adapter}: Adapters are compact neural networks that can be embedded between an LLM's layers. A well-known approach, adapter tuning \citep{houlsby2019parameter}, involves optimizing the adapter's parameters while leaving the original LLM parameters intact. These adapters are designed to adjust the intermediate layer representations to better align with task-specific needs.
2) \textbf{LLMs Themselves}: An alternative strategy involves modifying the LLM directly to produce task-specific representations by fine-tuning the model's weights across all or selected layers. This method of fine-tuning is prevalent for AE-LLMs \citep{jin2020bert} and has also been applied to AR-LLMs in early use of GPT-like models \citep{radford2018improving}.
3) \textbf{Output Layers}: Once task-specific representations are produced by either adapters or the LLM directly, the function of the output layers is to translate these representations into a designated output space. These layers typically consist of one or several linear layers. For example, linear functions are frequently used for tasks involving classification, while tasks that involve extractive question answering often necessitate the use of two linear functions to determine the beginning and concluding positions of the answer within a text passage.
4) \textbf{Activation Prefixes}: Within the scope of deploying LLMs via task-specific supervised learning, where training neural networks is common, prefix tuning \citep{li-liang-2021-prefix} presents an innovative method that employs prefixes to directly modify intermediate representations. These prefixes are essentially embeddings that are added at various layers, with dimensions identical to those of token embeddings, functioning as virtual tokens. Introducing these prefixes at earlier stages in the model allows for the infusion of task-specific information into more advanced layers, thereby improving the model's alignment with the desired task objectives.

Beyond the four channels previously outlined, verbal channels offer a unique approach for articulating the task context in which LLMs can identify and execute the intended tasks. These channels include:
5) \textbf{Verbal Free-form Context}:
In this approach, a context is articulated using free-form text, such as task instructions and few-shot demonstrations, which can activate complex cognitive functions. By merely incorporating task instructions within the context, AR-LLMs are enabled to undertake a multitude of tasks through zero-shot prompts. Another widely adopted method is few-shot prompting \citep{radford2019language,brown2020gpt3}, which involves learning from a limited number of examples for in-context learning without the need for gradient updates, showcasing a human-like efficiency in acquiring new tasks. This method is particularly effective in eliciting cognitive behaviors akin to those observed with few-shot demonstrations, with further details discussed in Section \ref{sec:think_plan_act}.
It's important to recognize that, in contrast to channels that are easily differentiated by input-side modalities (such as task-specific examples), this channel (e.g., task instructions) can intertwine with model inputs, e.g., task-specific examples. This allows for the seamless integration of the models' world knowledge into tasks, for instance, ``summarize deep learning technology''. 
6) \textbf{Contextual Text Patterns}:
Given their training on a denoising language model objective, AE-LLMs excel in completing texts by filling in missing words, a trait that can be leveraged for downstream tasks. Task-specific patterns, in this regard, serve as a mechanism to alter given task-specific examples. Typically, this involves appending the examples with a cloze-style phrase or sentence (text with missing words) tailored to the task, allowing the model to predict the intended task outcomes based on the placeholders filled within the text. Pattern Exploitation Training (PET) \citep{schick-schutze-2021-exploiting} involves the creative design of task-specific patterns and the fine-tuning of LLMs to these patterns. Conversely, auto-prompt methods \citep{shin-etal-2020-autoprompt} seek to optimize task-specific patterns to better fit the models, enhancing their ability to interpret and respond to the given tasks effectively.

\section{Evaluation of Modalities and Channels}
\label{sec:comparisons_of_paradigms}

\subsection{Evaluating Usability of Deployment Channels}
This section introduces a framework for assessing the usability of language model deployment channels, focusing on their customizability, transparency, and complexity, as summarized in Table \ref{tab:assess_channel}.

\paragraph{Customizability of User-level Tasks: Extent of User Control over Channels}
Essentially, any task can be articulated in human languages, such as English, using free-form context. This adaptability is a testament to the evolution of human language over thousands of years, which has been refined to describe a vast array of everyday and complex scientific problems. Typically, in a zero-shot learning context, the channel consists solely of task instructions within the prompts, capable of encompassing a wide range of tasks. For instance, Wang et al., \citet{wang-etal-2022-super} have converted standard NLP datasets designed for optimized channels into instruction-based formats for 76 different tasks.
Moreover, free-form task instructions allow for nuanced control mechanisms, including explicit directives (such as specifying output formats or initiating reasoning processes) and subtle cues (such as inducing cognitive behaviors through few-shot examples). These aspects will be further explored in Section \ref{sec:think_plan_act} and summarized in Table \ref{tab:think_plan_act}.
In contrast, since other channels are set during the optimization process for specific tasks, they lack the flexibility for user-directed modifications. Channels that require adjustments, such as fine-tuning the LLM, adapter tuning, or prefix tuning, rely on supervised learning methods for configuration. Although prompting in AE-LLMs could, in theory, facilitate task adjustments at inference time without prior task-specific fine-tuning—akin to AR-LLMs' prompting approach—it often requires task-specific optimization to achieve effective channel performance. For example, techniques like Pattern Exploitation Training (PET) \citep{schick-schutze-2021-exploiting} utilize mathematical optimization to adapt models to specific patterns, whereas Auto-prompt \citep{shin-etal-2020-autoprompt} optimizes text patterns for language models. The question of whether this need for optimization arises from the inherent complexities of auto-encoding language models invites further research. 

\paragraph{User-level Transparency: Can Channel Formulation Be Easily Understood by Users?}
The focus here is on the understandability of the channels themselves to lay users, rather than their functional effectiveness, as this greatly influences the user experience. For example, the objective of an output layer is clear --- transforming LLM representations into a specific output format. However, the process involving dense representations through matrix multiplication is not intuitively understandable to the non-specialist. Moreover, text patterns refined through AE-LLMs' Auto-prompting often lack the straightforwardness found in manually created prompts.

\paragraph{User-level Complexity: Assessing the Number of Conceptual Components}
This analysis evaluates the conceptual load required to deploy $T$ tasks using various channels, moving away from the parameter size metric, which is more pertinent to researchers and developers. Assuming each task is accommodable across all channels, we quantify the complexity as follows:
For fine-tuned LLMs, prefixes, adapters and output layers, each task-specific adjustment equates to a complexity of $T$, with 
$T$ denoting the total number of tasks. Additionally, $N$ text patterns are devised per task, resulting in a complexity of $N \times T$, where $N$ represents the number of patterns per task. The complexity for verbal free-form context is considered negligible, as these are formulated spontaneously by users at the time of use.
From this framework, we can deduce the complexity inherent to each deployment paradigm. For instance, LLM fine-tuning, which necessitates one LLM and one output layer per task, carries a complexity of $2 \times T$.

\subsection{Evaluating Expressiveness of Modalities}
During LLM fine-tuning and adapter tuning, the task-specific output layers strictly limit the range of possible outputs, hindering the potential for detailed expressiveness and, by extension, advanced cognitive behaviors. The output space is tightly defined, with actions or labels being pre-determined and given specific meanings through task-specific supervised learning. Nonetheless, certain probing techniques allow us to uncover the thought processes behind their predictions, a topic we will explore further in Section \ref{sec:think_fast_slow}.
When it comes to AE-LLMs prompted with text patterns, these models are limited to generating only specific tokens or words, constrained by the patterns set in advance. These constraints, such as token positions and quantities dictated by the input patterns, along with the need for grammatical and coherent text completion, restrict the models' ability to articulate complex ideas, plans, and actions.
On the other hand, AR-LLMs' prompting capitalizes on their auto-regressive nature to produce unbounded, free-form text, influenced solely by the given input context. This capability is further demonstrated in Section \ref{sec:think_plan_act} and summarized in Table \ref{tab:think_plan_act}, showcasing the open-ended expressiveness unique to the AR-LLM prompting paradigm.

\section{Cognitive Behaviors Under AR-LLMs' Prompting Paradigm}
\label{sec:think_plan_act}
This section elucidates the capability of AR-LLM prompting paradigms to exhibit cognitive behaviors expressed by the free-form modalities by mainpulating the free-form channels. It's important to clarify that not every AR-LLM demonstrates cognitive behaviors—smaller models like GPT-2 \citep{radford2019language} may not. 
Specifically, we analyze four cognitive behaviors: thinking, reasoning, planning, and feedback learning, leaving the examination of their interrelationships for future research.

\subsection{Thinking, Fast And Slow}
\label{sec:think_fast_slow}
At the core of cognitive behavior lies thinking. The Kahneman's framework \citep{kahneman2011thinking} divides thinking into two distinct systems: the fast system operates through intuitive shortcuts for quick navigation of daily situations without extensive analysis. Conversely, the slow system, or System 2, involves conscious, detailed and methodical examination of information, necessitating logical deliberation to arrive at decisions and address challenges.

\paragraph{Fast Thinking via Task-specific Channels}
Using channels trained through task-specific supervised learning can achieve performances that rival or exceed human performance.
Nonetheless, they often struggle with generalizing to data from natural domain shifts, adversarial perturbations and debiased data, as summarized by Li et al., \citet{li2023ood-eval}. This limitation is consistently attributed to shortcut learning, such as classifying sentences containing the word ``No'' as ``contradiction'' in text entailment tasks \citep{wallace2019universal,du-etal-2021-towards}.
The intriguing question arises whether task-specific channels can also develop System 2 --- the fast system.
While the limited expressiveness of task-specific outputs does not offer straightforward evidence, Li et al., \citet{li-liu-2023-make} employ a technical probe \citep{sundararajan2017axiomatic} to reveal that indulgence in shortcut learning during task-specific training impedes the development of the slow system. 
While the mentioned research primarily examines the LLM fine-tuning paradigm, it's our contention that shortcut learning and the fast thinking are likely prevalent across all the parametric channels, including prefixes and adapters, trained on supervised datasets to some degree. This is attributed to the inherent characteristics of gradient descent optimization, as demonstrated by empirical findings in Li et al., \citet{li-liu-2023-make}.
Another empirical evidence shows that methods like prefix and adapter tuning, although more resilient, still notably falter under distribution shifts and adversarial attacks \citep{han-etal-2021-robust,yang2022on}. 
The mitigated impact observed in prefix and adapter tuning is attributed to the fact that the underlying LLMs are not directly engaged as task-specific channels, as explored by \citep{han-etal-2021-robust}. While we draw parallels between reliance on shortcuts and fast thinking within human cognition, some research within the NLP field argues that such dependency on shortcuts (dataset biases) detracts from the models' relevance to human-level cognition \citep{zhong2023agieval}. This perspective arises from the view that the shortcuts might not reflect genuine human cognitive activities within the field of NLP.

\paragraph{Minimal Fast Thinking Evident with AR-LLMs Prompting}
Research findings \citep{si2023prompting,zhang-etal-2022-robustness} consistently indicate the difficulty of inducing fast thinking in AR-LLMs through prompting techniques. These models typically remain unfazed by various distributional shifts, such as domain shift and adversarial perturbations. Min et al., \citet{min-etal-2022-rethinking} demonstrate that, even with few-shot demonstrations for in-context learning, the models tend to leverage the structure of these demonstrations to organize the generation rather than relying on simplistic input-to-label mappings for predictions.
Additionally, Raman et al., \citet{raman-etal-2023-model} show that PET prompting improve the AE-LLMs' ability to withstand adversarial attacks. Nonetheless, this enhanced robustness is somewhat restricted. The constrained effectiveness could be attributed to the dependency on task-specific channels inherent during the deployment of the PET prompting.

\paragraph{Slow Thinking in Prompting Paradigms}
The remainder of this section will illustrate the capacity of AR-LLMs' prompting to replicate the human slow thinking process through the exhibition of effortful mental activities, as encapsulated in Table \ref{tab:think_plan_act}.

\subsection{Reasoning}
Reasoning is a thinking process to conclusions or decisions with the sequential and interconnected nature, i.e., chain-of-thoughts (CoTs) \citep{wei2022chain}. This is the most common definition in the NLP/LLM are to investigate the LLMs' reasoning ability. 
With a reasoning path in free-form modality, models can better solve complicated tasks requiring multi-step reasoning compared to the conclusion without CoTs. As an illustration, Wei et al., \citet{wei2022chain} substantially boosts model efficacy in solving mathematical reasoning bechmarks.

Reasoning is defined as the process of arriving at conclusions or decisions through a sequential and interconnected series of thoughts, often referred to as a chain-of-thoughts (CoTs) \citep{wei2022chain}. This definition is widely accepted in the field of Natural Language Processing (NLP) for exploring the reasoning capabilities of LLMs. By employing a reasoning path via the modality of free-form text, models are more adept at tackling complex tasks that necessitate multi-step reasoning, as opposed to reaching conclusions without the aid of CoTs.
Technically, the auto-regressive nature employs the thoughts or intermediate steps generated as the prior for generating subsequent thoughts and, ultimately, the final predictions.

\paragraph{Context for Eliciting Reasoning}
Two primary contexts are employed to facilitate the creation of intermediate reasoning steps: incorporating a Chain of Thought (CoT) triggers in task instructions (\textbf{zero-shot CoTs}), such as ``Let's think step-by-step'' \citep{kojima2022large}, within prompts, or integrating manually crafted reasoning steps in a few-shot learning context (\textbf{few-shot CoTs}) \citep{wei2022chain}. To circumvent the manual compilation of few-shot demonstrations with reasoning sequences, Zhang et al., 
\citet{zhang2023automatic} developed a method to automatically generate few-shot demonstrations by choosing several queries and utilizing zero-shot CoTs to craft reasoning sequences for each query (\textbf{Auto CoTs}).
Given that simple greedy decoding (producing a single chain) is prone to error accumulation in intermediate steps, Wang et al., 
\citet{wang2023selfconsistency}
propose generating multiple chains and consolidating them through majority voting, thereby enhancing model accuracy in both scenarios (\textbf{CoTs-SC}).

\subsection{Planning}
Planning involves the forethought and organization of actions or steps to achieve a predetermined objective. This process fundamentally requires a comprehension or representation of the environment and involves breaking down tasks into smaller, manageable subgoals. 
It represents a key cognitive behavior modeled within the fields of AI. Typical planning methods break down tasks into subgoals through explicit symbolic representation \citep{russell2010artificial}. For instance, partial-order planning ensures the logical sequencing of actions by modeling actions, preconditions, effects, and the relations among actions in such a way that actions are logically sequenced to meet the goal's preconditions.
Differing from traditional approaches that rely on explicitly modeled knowledge and reasoning mechanisms, LLMs leverage their inherent knowledge and inferential capabilities to mimic planning. They do this by producing text sequences that suggest a logical progression of steps or actions directed towards an objective \citep{hao-etal-2023-reasoning,wang-etal-2023-plan,huang2022language}. This skill stems from the models' proficiency in forecasting the subsequent most likely word sequence based on a context indicative of planning or reasoning processes.

\paragraph{Context to Elicit Plans}
Similar to the activation of reasoning processes, the process of planning can be prompted through the inclusion of specific planning cues in zero-shot scenarios, such as the prompt ``let's carry out the plan'' \citep{wang-etal-2023-plan}, or through the demonstration of planning steps in few-shot examples \citep{huang2022language}. Experimental findings indicate that instructions tailored to tasks significantly enhance the performance of LLMs on various tasks. For instance, directives like ``pay attention to calculation'' \citep{hao-etal-2023-reasoning} or ``identify key variables and their corresponding figures to formulate a plan'' \citep{wang-etal-2023-plan} have been shown to improve outcomes in tasks requiring numerical reasoning.

\paragraph{Applying Planning for Sequential Decision-making}
This ability is essential for addressing problems requiring a series of decisions, especially when deploying LLMs in open-world scenarios like robotics. In such environments, tasks typically need physical actions (grounded), involve translating broad objectives into actionable steps (high-level), and present a vast range of possible actions (open-ended). Research has demonstrated the effectiveness of LLMs in deconstructing complex goals into actionable sequences within such dynamic environments, as seen in projects like ALFWorld \citep{yao2023react}, VirtualHome \citep{huang2022language}, and Minecraft \citep{wang2023describe}.
An example from ALFWorld illustrates this: achieving the objective of ``examining paper under desklamp'' necessitates LLMs to devise practical plans (e.g., initially approaching the coffee table, then acquiring the paper and utilizing the desklamp) and subsequently generate textual instructions for execution in real-world settings.

\subsection{Feedback Learning}
As Kahneman et al., \citet{kahneman2011thinking} elucidates, although System 1 may rush to judgments that are biased or erroneous, System 2 has the capacity to identify and rectify these mistakes through introspection on the rapid decisions made by System 1. Similarly, LLMs have shown the ability to mimic this aspect of human cognition.

\paragraph{Feedback Generation and Contextual Basis}
LLMs are adept at generating feedback by reflecting on their previously given responses or observations from interactions with the external environment. There are two primary scenarios for feedback generation:
1) Feedback based on previous actions and external feedback: In the work by Yao et al., \citet{yao2023react}, LLMs engaging with a Wikipedia API to search for entities that do not exist, such as ``Search[goddess frigg]'', may encounter a 404 error, delivered in JSON format. In response, LLMs can articulate feedback about the error related to their action, such as stating, ``Could not find goddess frigg.''.
2) Feedback based solely on prior responses: 
This approach is relevant in various situations where external environmental feedback is absent \citep{shinn2023reflexion,hao-etal-2023-reasoning}. In such cases, LLMs can give feedback on previous answers by applying certain evaluation metrics, such as determining the relevance of a sub-question to a broader question requiring intricate, multi-step reasoning \citep{hao-etal-2023-reasoning}. This process involves using a prompt that asks the LLM to judge the utility of a sub-question in addressing the main question (Prompt: ``Given a question, assess if the subquestion aids in solving the original question. Answer 'Yes' or 'No'. Question: \{goal\}; Subquestion: \{action\}. Is the subquestion useful?'').
Furthermore, feedback may be presented as numerical scores, such as the confidence scores (normalized logits) for 'Yes' or 'No' answers, instead of in verbal form. The decision to use numerical rather than verbal feedback is contingent on the specific requirements of the feedback mechanism, as explored in subsequent discussions.

\paragraph{Mechanism for Feedback Learning}
After generating feedback, Shinn et al., \citet{shinn2023reflexion} directly include verbal feedback LLMs have produced to enhance the accuracy of their responses or decisions. 
Meanwhile, Hao et al., \citet{hao-etal-2023-reasoning} detail a methodology where numerical feedback serves as a reward system for guiding LLMs for action selection. Following this, the LLMs function as world models to predict the subsequent state of state-action pairs during the planning phase, utilizing Monte-Carlo Tree Search (MCTS) to achieve this aim. Instead of allowing LLMs to directly process the feedback, an implicit feedback learning strategy is employed where a feedback loop is deliberately established to influence the sequence of actions undertaken by the LLMs via the update of state values during the propagation phase of MCTS.


\begin{table}[ht!]
    \centering
    \footnotesize
    \begin{tabular}{p{1.3cm}p{2.5cm}p{3.5cm}}
        \toprule
         Behaviors &  Context  & Relevant Works \\
        \midrule
        \multirow{2}{*}{Reasoning}  & CoT triggers, e.g., ``Let's think step by step.'' & Zero-shot CoTs \citep{kojima2022large}, \newline 
        Auto-CoTs \citep{zhang2023automatic} 
        \\ \cmidrule{2-3}
        
        &  Few-shot demos with CoTs & Few-shot CoTs \citep{wei2022chain}, 
        \newline 
        CoTs-SC \citep{wang2023selfconsistency}, 
        \newline 
        Auto-prompt \citep{zhang2023automatic},
        \newline 
        ToT \citep{yao2023tree}
        \\ \hline 
        
        \multirow{2}{*}{Planning} & Zero-shot instruction & Wang et al., \citet{wang-etal-2023-plan} 
        \\ \cmidrule{2-3}
        & Few-shot demos with planning steps & Huang et al., \citet{huang2022language}
        \\ \hline
        
        Feedback \newline Learning &  Observations from external environments  & Reflexion \citep{shinn2023reflexion}
        \\ \cmidrule{2-3}
        
         & Previous answers/decisions &  Self-refine \citep{madaan2023selfrefine}, 
         \newline 
         Reflexion 
         \citep{shinn2023reflexion}, 
        \newline 
        RAP \citep{hao-etal-2023-reasoning} 
        \\ 
        \bottomrule
    \end{tabular}
    
    \caption{Cognitive behaviors enabled by free-form context. For the ``Feedback Learning'' sections, we illustrate the contexts utilized to produce feedback. It's worth noting that the methods for feedback adaptation might not always employ free-form context; for instance, they may involve advanced search techniques as outlined in our study. The final column presents examples of tasks for demonstration purposes, though the list is not comprehensive.}
    \label{tab:think_plan_act}
\end{table}

\section{Bridging LLM Deployment Gaps with Insights from Cognitive Behaviors}
\label{sec:challenges}
This section investigates how insights into cognitive behaviors can aid in addressing the tuning and deployment challenges faced by LLMs operating as autonomous agents and within multi-agent systems.

\paragraph{Autonomous Cognitive Behaviors}
Instead of relying on explicit contextual cues to trigger advanced cognitive functions, an intelligent system is expected to independently engage in reasoning, planning, and decision-making as it interacts with the external world---for instance, by seeking input from humans or utilizing available tools. To foster such autonomous behaviors, various algorithms aim to tune LLMs for independently exhibiting behaviors that align with human cognitive processes. For instance, Liu et al., \citet{liu2023logicot} have developed techniques for instruction tuning that facilitates autonomous reasoning.
Yet, the challenge remains in creating instructional data that encapsulates higher-order cognitive functions. A pivotal question emerges:
\emph{How can various cognitive behaviors be encapsulated within free-form text (instruction data)?}  Addressing this question is crucial for ensuring that the data used for tuning mirrors human cognitive processes, thereby making the resulting model actions more human-like. Unraveling this issue might necessitate insights from both cognitive psychology and linguistics.
Another approach to tuning involves the use of reliable reward models, such as reinforcement learning from human feedback (RLHF) \citep{ouyang2022training} and behavior cloning \citep{nakano2021webgpt}.
Many studies \citep{ouyang2022training,nakano2021webgpt} develop reward models based on comparisons of model-generated responses, with human evaluators ranking these responses.
An unresolved inquiry remains:
\emph{How can reward models be devised to truly reflect human cognitive preferences?}

\paragraph{Navigating Free-form Contexts in Multi-turn Interactions Within Multi-agent Systems}
In exploring planning and feedback learning strategies, it becomes evident that multi-agent systems are designed to facilitate interactions between LLMs and the external world, as well as among LLMs themselves. Examples include LLMs acting as both evaluators and actors in feedback learning environments \citep{shinn2023reflexion} or taking on roles as evaluators, actors, and environmental simulators in planning scenarios \citep{hao-etal-2023-reasoning}.
Such setups necessitate more than a one-time interaction between LLMs and users, requiring instead sustained, multi-turn dialogues.
Challenges emerge within these complex interactions: \emph{How do agents process and integrate information from other agents and their own previous dialogues? How can environmental data be stringified into a format understandable by LLMs? 
What strategies can simplify the management of extended sequences of interaction trajectories?}

\paragraph{Advancing Towards A Unified Inference Framework for Multi-agent Systems}
Despite the recent development of various LLM-based multi-agent systems, there remains an absence of a unified framework across these models. 
Exploring such framework serves as a key drive for this research on a cognitive framework. 
Organizing LLM-based agents by their cognitive behaviors offers a pathway to this unification. For instance, as indicated in Table \ref{tab:think_plan_act}, the ReAct framework \citep{yao2023react} employs a reasoning agent for decision-making, whereas the RAP framework \citep{hao-etal-2023-reasoning} utilizes a reasoning agent for decision-making alongside a feedback learning agent for evaluation.
Delving deeper into the relationships between cognitive behaviors might benefit from insights in cognitive psychology. Taking the concept of Self-Regulated Learning (SRL) as defined by \citet{zimmerman2000attaining}, planning and feedback learning are intertwined, enhancing the learning process.

``Self-regulated learning refers to self-generated thoughts, feelings, and actions that are planned and cyclically adapted to the attainment of personal goals''

This rationale lays the groundwork for future efforts to combine planning frameworks (like RAP) with feedback learning frameworks (such as Reflexion).
Furthermore, the structure of certain inference models showcases their interconnections and the presence of common detailed components. 
For instance, both feedback learning frameworks, exemplified by Reflexion \citet{shinn2023reflexion}, and planning frameworks, such as RAP \citet{hao-etal-2023-reasoning}, incorporate a common detailed module: a feedback learning agent. This agent plays a pivotal role in decision-making by facilitating the selection of appropriate actions from an assortment of tools and environments. 
It is evident that the interconnection within the intricate inference framework can be examined, especially at the level of cognitive behavior.
Our comprehensive analysis aims to pave the way for the systematic creation of multi-agent LLM-based frameworks, or inversely, to stimulate research in cognitive psychology.
 
\section{Conclusion}
In summary, our analysis seeks to inspire further research in AI, within the domain of language intelligence and beyond, to move away from heavily optimized task-specific channels. Instead, we advocate for the adoption of natural and free-form modalities throughout the pretraining phase via self-supervised learning, followed by straightforward inference-time deployment that eschews the necessity for mathematically optimizing task-specific channels. 
We developed an analytical framework to examine the deployment of LLMs to reach the conclusion.
Besides, the auto-regressive nature of free-form modalities, leveraged during pretraining, enhances the capacity for exhibiting a range of human-like cognitive behaviors by utilizing the free-form channel. It is important to clarify that we do not advocate that LLMs possess conscious thought. Rather, our findings illustrate how LLMs, such as ChatGPT, can imitate the outcomes of human cognitive activities via the free-form modality given suitable verbal context.
Lastly, we highlight the opportunity to address challenges in LLM deployment through the integration of cognitive behavior concepts.

\bibliographystyle{named}
\bibliography{ijcai24}

\begin{thebibliography}{}

\bibitem[\protect\citeauthoryear{Bartneck \bgroup \em et al.\egroup }{2020}]{bartneck2020human}
Christoph Bartneck, Tony Belpaeme, Friederike Eyssel, Takayuki Kanda, Merel Keijsers, and Selma {\v{S}}abanovi{\'c}.
\newblock {\em Human-robot interaction: An introduction}.
\newblock Cambridge University Press, 2020.

\bibitem[\protect\citeauthoryear{Bengio \bgroup \em et al.\egroup }{2003}]{bengio-2003-neural-lm}
Yoshua Bengio, R\'{e}jean Ducharme, Pascal Vincent, and Christian Janvin.
\newblock A neural probabilistic language model.
\newblock {\em JMLR}, 3:1137–1155, 2003.

\bibitem[\protect\citeauthoryear{Brown \bgroup \em et al.\egroup }{2020}]{brown2020gpt3}
Tom Brown, Benjamin Mann, Nick Ryder, Melanie Subbiah, Jared~D Kaplan, Prafulla Dhariwal, Arvind Neelakantan, Pranav Shyam, Girish Sastry, Amanda Askell, Sandhini Agarwal, Ariel Herbert-Voss, Gretchen Krueger, Tom Henighan, Rewon Child, Aditya Ramesh, Daniel Ziegler, Jeffrey Wu, Clemens Winter, Chris Hesse, Mark Chen, Eric Sigler, Mateusz Litwin, Scott Gray, Benjamin Chess, Jack Clark, Christopher Berner, Sam McCandlish, Alec Radford, Ilya Sutskever, and Dario Amodei.
\newblock Language models are few-shot learners.
\newblock In {\em Advances in NIPS}, volume~33, pages 1877--1901, 2020.

\bibitem[\protect\citeauthoryear{Devlin \bgroup \em et al.\egroup }{2019}]{devlin-bert2018}
Jacob Devlin, Ming-Wei Chang, Kenton Lee, and Kristina Toutanova.
\newblock {BERT}: Pre-training of deep bidirectional transformers for language understanding.
\newblock In {\em NAACL-HLT}, 2019.

\bibitem[\protect\citeauthoryear{Du \bgroup \em et al.\egroup }{2021}]{du-etal-2021-towards}
Mengnan Du, Varun Manjunatha, Rajiv Jain, Ruchi Deshpande, Franck Dernoncourt, Jiuxiang Gu, Tong Sun, and Xia Hu.
\newblock Towards interpreting and mitigating shortcut learning behavior of {NLU} models.
\newblock In {\em NAACL-HLT}, pages 915--929, 2021.

\bibitem[\protect\citeauthoryear{Han \bgroup \em et al.\egroup }{2021}]{han-etal-2021-robust}
Wenjuan Han, Bo~Pang, and Ying~Nian Wu.
\newblock Robust transfer learning with pretrained language models through adapters.
\newblock In {\em ACL-IJCNLP}, pages 854--861, 2021.

\bibitem[\protect\citeauthoryear{Hao \bgroup \em et al.\egroup }{2023}]{hao-etal-2023-reasoning}
Shibo Hao, Yi~Gu, Haodi Ma, Joshua Hong, Zhen Wang, Daisy Wang, and Zhiting Hu.
\newblock Reasoning with language model is planning with world model.
\newblock In {\em EMNLP}, pages 8154--8173, 2023.

\bibitem[\protect\citeauthoryear{Houlsby \bgroup \em et al.\egroup }{2019}]{houlsby2019parameter}
Neil Houlsby, Andrei Giurgiu, Stanislaw Jastrzebski, Bruna Morrone, Quentin De~Laroussilhe, Andrea Gesmundo, Mona Attariyan, and Sylvain Gelly.
\newblock Parameter-efficient transfer learning for nlp.
\newblock In {\em ICML}, pages 2790--2799, 2019.

\bibitem[\protect\citeauthoryear{Huang \bgroup \em et al.\egroup }{2022}]{huang2022language}
Wenlong Huang, Pieter Abbeel, Deepak Pathak, and Igor Mordatch.
\newblock Language models as zero-shot planners: Extracting actionable knowledge for embodied agents.
\newblock In {\em ICML}, pages 9118--9147, 2022.

\bibitem[\protect\citeauthoryear{Jin \bgroup \em et al.\egroup }{2020}]{jin2020bert}
Di~Jin, Zhijing Jin, Joey~Tianyi Zhou, and Peter Szolovits.
\newblock Is bert really robust? a strong baseline for natural language attack on text classification and entailment.
\newblock In {\em AAAI}, volume~34, pages 8018--8025, 2020.

\bibitem[\protect\citeauthoryear{Joshi \bgroup \em et al.\egroup }{2020}]{joshi-etal-2020-spanbert}
Mandar Joshi, Danqi Chen, Yinhan Liu, Daniel~S. Weld, Luke Zettlemoyer, and Omer Levy.
\newblock {S}pan{BERT}: Improving pre-training by representing and predicting spans.
\newblock {\em TACL}, 8:64--77, 2020.

\bibitem[\protect\citeauthoryear{Kahneman}{2011}]{kahneman2011thinking}
D.~Kahneman.
\newblock {\em Thinking, Fast and Slow}.
\newblock Farrar, Straus and Giroux, 2011.

\bibitem[\protect\citeauthoryear{Kojima \bgroup \em et al.\egroup }{2022}]{kojima2022large}
Takeshi Kojima, Shixiang~Shane Gu, Machel Reid, Yutaka Matsuo, and Yusuke Iwasawa.
\newblock Large language models are zero-shot reasoners.
\newblock In {\em Advances in NIPS}, 2022.

\bibitem[\protect\citeauthoryear{Lewis \bgroup \em et al.\egroup }{2020}]{lewis-etal-2020-bart}
Mike Lewis, Yinhan Liu, Naman Goyal, Marjan Ghazvininejad, Abdelrahman Mohamed, Omer Levy, Veselin Stoyanov, and Luke Zettlemoyer.
\newblock {BART}: Denoising sequence-to-sequence pre-training for natural language generation, translation, and comprehension.
\newblock In {\em ACL}, pages 7871--7880, 2020.

\bibitem[\protect\citeauthoryear{Li and Liang}{2021}]{li-liang-2021-prefix}
Xiang~Lisa Li and Percy Liang.
\newblock Prefix-tuning: Optimizing continuous prompts for generation.
\newblock In Chengqing Zong, Fei Xia, Wenjie Li, and Roberto Navigli, editors, {\em ACL-IJCNLP}, 2021.

\bibitem[\protect\citeauthoryear{Li and Liu}{2023}]{li-liu-2023-make}
Xinzhe Li and Ming Liu.
\newblock Make text unlearnable: Exploiting effective patterns to protect personal data.
\newblock In {\em TrustNLP}, pages 249--259, 2023.

\bibitem[\protect\citeauthoryear{Li \bgroup \em et al.\egroup }{2023}]{li2023ood-eval}
Xinzhe Li, Ming Liu, Shang Gao, and Wray Buntine.
\newblock A survey on out-of-distribution evaluation of neural nlp models.
\newblock In {\em IJCAI-23}, 7 2023.

\bibitem[\protect\citeauthoryear{Liu \bgroup \em et al.\egroup }{2019}]{liu2019roberta}
Yinhan Liu, Myle Ott, Naman Goyal, Jingfei Du, Mandar Joshi, Danqi Chen, Omer Levy, Mike Lewis, Luke Zettlemoyer, and Veselin Stoyanov.
\newblock Roberta: {A} robustly optimized {BERT} pretraining approach.
\newblock {\em CoRR}, abs/1907.11692, 2019.

\bibitem[\protect\citeauthoryear{Liu \bgroup \em et al.\egroup }{2023}]{liu2023logicot}
Hanmeng Liu, Zhiyang Teng, Leyang Cui, Chaoli Zhang, Qiji Zhou, and Yue Zhang.
\newblock Logicot: Logical chain-of-thought instruction tuning.
\newblock In {\em EMNLP}, 2023.

\bibitem[\protect\citeauthoryear{Madaan \bgroup \em et al.\egroup }{2023}]{madaan2023selfrefine}
Aman Madaan, Niket Tandon, Prakhar Gupta, Skyler Hallinan, Luyu Gao, Sarah Wiegreffe, Uri Alon, Nouha Dziri, Shrimai Prabhumoye, Yiming Yang, Shashank Gupta, Bodhisattwa~Prasad Majumder, Katherine Hermann, Sean Welleck, Amir Yazdanbakhsh, and Peter Clark.
\newblock Self-refine: Iterative refinement with self-feedback.
\newblock In {\em Advances in NIPS}, 2023.

\bibitem[\protect\citeauthoryear{Min \bgroup \em et al.\egroup }{2022}]{min-etal-2022-rethinking}
Sewon Min, Xinxi Lyu, Ari Holtzman, Mikel Artetxe, Mike Lewis, Hannaneh Hajishirzi, and Luke Zettlemoyer.
\newblock Rethinking the role of demonstrations: What makes in-context learning work?
\newblock In {\em EMNLP}, pages 11048--11064, 2022.

\bibitem[\protect\citeauthoryear{Nakano \bgroup \em et al.\egroup }{2021}]{nakano2021webgpt}
Reiichiro Nakano, Jacob Hilton, Suchir Balaji, Jeff Wu, Long Ouyang, Christina Kim, Christopher Hesse, Shantanu Jain, Vineet Kosaraju, William Saunders, et~al.
\newblock Webgpt: Browser-assisted question-answering with human feedback.
\newblock {\em arXiv preprint arXiv:2112.09332}, 2021.

\bibitem[\protect\citeauthoryear{Ouyang \bgroup \em et al.\egroup }{2022}]{ouyang2022training}
Long Ouyang, Jeffrey Wu, Xu~Jiang, Diogo Almeida, Carroll Wainwright, Pamela Mishkin, Chong Zhang, Sandhini Agarwal, Katarina Slama, Alex Gray, John Schulman, Jacob Hilton, Fraser Kelton, Luke Miller, Maddie Simens, Amanda Askell, Peter Welinder, Paul Christiano, Jan Leike, and Ryan Lowe.
\newblock Training language models to follow instructions with human feedback.
\newblock In {\em Advances in NIPS}, 2022.

\bibitem[\protect\citeauthoryear{Radford \bgroup \em et al.\egroup }{2018}]{radford2018improving}
Alec Radford, Karthik Narasimhan, Tim Salimans, Ilya Sutskever, et~al.
\newblock Improving language understanding by generative pre-training.
\newblock 2018.

\bibitem[\protect\citeauthoryear{Radford \bgroup \em et al.\egroup }{2019}]{radford2019language}
Alec Radford, Jeff Wu, Rewon Child, David Luan, Dario Amodei, and Ilya Sutskever.
\newblock Language models are unsupervised multitask learners.
\newblock 2019.

\bibitem[\protect\citeauthoryear{Raman \bgroup \em et al.\egroup }{2023}]{raman-etal-2023-model}
Mrigank Raman, Pratyush Maini, J~Kolter, Zachary Lipton, and Danish Pruthi.
\newblock Model-tuning via prompts makes {NLP} models adversarially robust.
\newblock In {\em EMNLP}, pages 9266--9286, 2023.

\bibitem[\protect\citeauthoryear{Russell and Norvig}{2010}]{russell2010artificial}
Stuart~J Russell and Peter Norvig.
\newblock {\em Artificial intelligence a modern approach}.
\newblock London, 2010.

\bibitem[\protect\citeauthoryear{Schick and Sch{\"u}tze}{2021}]{schick-schutze-2021-exploiting}
Timo Schick and Hinrich Sch{\"u}tze.
\newblock Exploiting cloze-questions for few-shot text classification and natural language inference.
\newblock In {\em EACL}, pages 255--269, 2021.

\bibitem[\protect\citeauthoryear{Shin \bgroup \em et al.\egroup }{2020}]{shin-etal-2020-autoprompt}
Taylor Shin, Yasaman Razeghi, Robert~L. Logan~IV, Eric Wallace, and Sameer Singh.
\newblock {A}uto{P}rompt: {E}liciting {K}nowledge from {L}anguage {M}odels with {A}utomatically {G}enerated {P}rompts.
\newblock In {\em EMNLP}, pages 4222--4235, 2020.

\bibitem[\protect\citeauthoryear{Shinn \bgroup \em et al.\egroup }{2023}]{shinn2023reflexion}
Noah Shinn, Federico Cassano, Ashwin Gopinath, Karthik~R Narasimhan, and Shunyu Yao.
\newblock Reflexion: language agents with verbal reinforcement learning.
\newblock In {\em Advances in NIPS}, 2023.

\bibitem[\protect\citeauthoryear{Si \bgroup \em et al.\egroup }{2023}]{si2023prompting}
Chenglei Si, Zhe Gan, Zhengyuan Yang, Shuohang Wang, Jianfeng Wang, Jordan~Lee Boyd-Graber, and Lijuan Wang.
\newblock Prompting {GPT}-3 to be reliable.
\newblock In {\em ICLR}, 2023.

\bibitem[\protect\citeauthoryear{Sundararajan \bgroup \em et al.\egroup }{2017}]{sundararajan2017axiomatic}
Mukund Sundararajan, Ankur Taly, and Qiqi Yan.
\newblock Axiomatic attribution for deep networks.
\newblock In {\em ICML}, pages 3319--3328, 2017.

\bibitem[\protect\citeauthoryear{Wallace \bgroup \em et al.\egroup }{2019}]{wallace2019universal}
Eric Wallace, Shi Feng, Nikhil Kandpal, Matt Gardner, and Sameer Singh.
\newblock Universal adversarial triggers for attacking and analyzing nlp.
\newblock In {\em EMNLP-IJCNLP}, pages 2153--2162, 2019.

\bibitem[\protect\citeauthoryear{Wang \bgroup \em et al.\egroup }{2019}]{wang2018glue}
Alex Wang, Amanpreet Singh, Julian Michael, Felix Hill, Omer Levy, and Samuel~R. Bowman.
\newblock {GLUE:} {A} multi-task benchmark and analysis platform for natural language understanding.
\newblock In {\em ICLR}. OpenReview.net, 2019.

\bibitem[\protect\citeauthoryear{Wang \bgroup \em et al.\egroup }{2022}]{wang-etal-2022-super}
Yizhong Wang, Swaroop Mishra, Pegah Alipoormolabashi, Yeganeh Kordi, Amirreza Mirzaei, Atharva Naik, Arjun Ashok, Arut~Selvan Dhanasekaran, Anjana Arunkumar, David Stap, Eshaan Pathak, Giannis Karamanolakis, Haizhi Lai, Ishan Purohit, Ishani Mondal, Jacob Anderson, Kirby Kuznia, Krima Doshi, Kuntal~Kumar Pal, Maitreya Patel, Mehrad Moradshahi, Mihir Parmar, Mirali Purohit, Neeraj Varshney, Phani~Rohitha Kaza, Pulkit Verma, Ravsehaj~Singh Puri, Rushang Karia, Savan Doshi, Shailaja~Keyur Sampat, Siddhartha Mishra, Sujan Reddy~A, Sumanta Patro, Tanay Dixit, and Xudong Shen.
\newblock Super-{N}atural{I}nstructions: Generalization via declarative instructions on 1600+ {NLP} tasks.
\newblock In {\em EMNLP}, pages 5085--5109, 2022.

\bibitem[\protect\citeauthoryear{Wang \bgroup \em et al.\egroup }{2023a}]{wang-etal-2023-plan}
Lei Wang, Wanyu Xu, Yihuai Lan, Zhiqiang Hu, Yunshi Lan, Roy Ka-Wei Lee, and Ee-Peng Lim.
\newblock Plan-and-solve prompting: Improving zero-shot chain-of-thought reasoning by large language models.
\newblock In {\em ACL}, pages 2609--2634, 2023.

\bibitem[\protect\citeauthoryear{Wang \bgroup \em et al.\egroup }{2023b}]{wang2023selfconsistency}
Xuezhi Wang, Jason Wei, Dale Schuurmans, Quoc~V Le, Ed~H. Chi, Sharan Narang, Aakanksha Chowdhery, and Denny Zhou.
\newblock Self-consistency improves chain of thought reasoning in language models.
\newblock In {\em ICLR}, 2023.

\bibitem[\protect\citeauthoryear{Wang \bgroup \em et al.\egroup }{2023c}]{wang2023describe}
Zihao Wang, Shaofei Cai, Guanzhou Chen, Anji Liu, Xiaojian Ma, and Yitao Liang.
\newblock Describe, explain, plan and select: Interactive planning with {LLM}s enables open-world multi-task agents.
\newblock In {\em Advances in NIPS}, 2023.

\bibitem[\protect\citeauthoryear{Wei \bgroup \em et al.\egroup }{2022a}]{wei2022emergent}
Jason Wei, Yi~Tay, Rishi Bommasani, Colin Raffel, Barret Zoph, Sebastian Borgeaud, Dani Yogatama, Maarten Bosma, Denny Zhou, Donald Metzler, Ed~H. Chi, Tatsunori Hashimoto, Oriol Vinyals, Percy Liang, Jeff Dean, and William Fedus.
\newblock Emergent abilities of large language models.
\newblock {\em TMLR}, 2022.

\bibitem[\protect\citeauthoryear{Wei \bgroup \em et al.\egroup }{2022b}]{wei2022chain}
Jason Wei, Xuezhi Wang, Dale Schuurmans, Maarten Bosma, brian ichter, Fei Xia, Ed~H. Chi, Quoc~V Le, and Denny Zhou.
\newblock Chain of thought prompting elicits reasoning in large language models.
\newblock In {\em Advances in NIPS}, 2022.

\bibitem[\protect\citeauthoryear{Yang and Liu}{2022}]{yang2022on}
Zonghan Yang and Yang Liu.
\newblock On robust prefix-tuning for text classification.
\newblock In {\em ICLR}, 2022.

\bibitem[\protect\citeauthoryear{Yao \bgroup \em et al.\egroup }{2023a}]{yao2023tree}
Shunyu Yao, Dian Yu, Jeffrey Zhao, Izhak Shafran, Thomas~L. Griffiths, Yuan Cao, and Karthik~R Narasimhan.
\newblock Tree of thoughts: Deliberate problem solving with large language models.
\newblock In {\em Advances in NIPS}, 2023.

\bibitem[\protect\citeauthoryear{Yao \bgroup \em et al.\egroup }{2023b}]{yao2023react}
Shunyu Yao, Jeffrey Zhao, Dian Yu, Nan Du, Izhak Shafran, Karthik~R Narasimhan, and Yuan Cao.
\newblock React: Synergizing reasoning and acting in language models.
\newblock In {\em ICLR}, 2023.

\bibitem[\protect\citeauthoryear{Zhang \bgroup \em et al.\egroup }{2022}]{zhang-etal-2022-robustness}
Hongxin Zhang, Yanzhe Zhang, Ruiyi Zhang, and Diyi Yang.
\newblock Robustness of demonstration-based learning under limited data scenario.
\newblock In {\em EMNLP}, pages 1769--1782, 2022.

\bibitem[\protect\citeauthoryear{Zhang \bgroup \em et al.\egroup }{2023}]{zhang2023automatic}
Zhuosheng Zhang, Aston Zhang, Mu~Li, and Alex Smola.
\newblock Automatic chain of thought prompting in large language models.
\newblock In {\em ICLR}, 2023.

\bibitem[\protect\citeauthoryear{Zhong \bgroup \em et al.\egroup }{2023}]{zhong2023agieval}
Wanjun Zhong, Ruixiang Cui, Yiduo Guo, Yaobo Liang, Shuai Lu, Yanlin Wang, Amin Saied, Weizhu Chen, and Nan Duan.
\newblock Agieval: A human-centric benchmark for evaluating foundation models.
\newblock {\em arXiv preprint arXiv:2304.06364}, 2023.

\bibitem[\protect\citeauthoryear{Zimmerman}{2000}]{zimmerman2000attaining}
Barry~J Zimmerman.
\newblock Attaining self-regulation: A social cognitive perspective.
\newblock In {\em Handbook of self-regulation}, pages 13--39. 2000.

\end{thebibliography}

\end{document}